\documentclass[10pt,twocolumn,letterpaper]{article}
\usepackage[accsupp]{axessibility}
\usepackage{iccv}
\usepackage{times}
\usepackage{epsfig}
\usepackage{graphicx}
\usepackage{amsmath}
\usepackage{amssymb}
\usepackage{authblk}
\usepackage{amsfonts,amsthm}
\usepackage{array}
\usepackage{balance}
\usepackage{booktabs}
\usepackage{color}
\usepackage{enumitem}
\usepackage{float}
\usepackage{mathtools}
\usepackage{multirow}
\usepackage{setspace}
\usepackage{subcaption}
\usepackage{url}

\theoremstyle{definition}

\newcolumntype{L}[1]{>{\raggedright\let\newline\\\arraybackslash\hspace{0pt}}m{#1}}
\newcolumntype{C}[1]{>{\centering\let\newline\\\arraybackslash\hspace{0pt}}m{#1}}
\newcolumntype{R}[1]{>{\raggedleft\let\newline\\\arraybackslash\hspace{0pt}}m{#1}}

\setlist[itemize]{noitemsep, topsep=0pt}
\setlist[enumerate]{noitemsep, topsep=0pt}

\newcommand{\parens}[1]{\left(#1\right)}

\newcommand{\norm}[1]{\left\Vert#1\right\Vert}

\usepackage[breaklinks=true,bookmarks=false]{hyperref}

\iccvfinalcopy 

\def\iccvPaperID{7714} 

\ificcvfinal\fi

\begin{document}

\title{Synthesis of Compositional Animations from Textual Descriptions}

\author[1,3]{Anindita Ghosh }
\author[1,2,3]{Noshaba Cheema \thanks{Supervising Author}}
\author[1,3]{Cennet Oguz}
\author[2,3]{Christian Theobalt}
\author[1,3]{Philipp Slusallek}
\affil[1]{German Research Center for Artificial Intelligence (DFKI)}
\affil[2]{Max-Planck Institute for Informatics}
\affil[3]{Saarland Informatics Campus}
\maketitle
\ificcvfinal\thispagestyle{empty}\fi


\begin{abstract}
    ``How can we animate 3D-characters from a movie script or move robots by simply telling them what we would like them to do?" ``How unstructured and complex can we make a sentence and still generate plausible movements from it?" These are questions that need to be answered in the long-run, as the field is still in its infancy.
    Inspired by these problems, we present a new technique for generating compositional actions, which handles complex input sentences.
    Our output is a 3D pose sequence depicting the actions in the input sentence. We propose a hierarchical two-stream sequential model to explore a finer joint-level mapping between natural language sentences and 3D pose sequences corresponding to the given motion. We learn two manifold representations of the motion -- one each for the upper body and the lower body movements. Our model can generate plausible pose sequences for short sentences describing single actions as well as long compositional sentences describing multiple sequential and superimposed actions.
    We evaluate our proposed model on the publicly available KIT Motion-Language Dataset containing 3D pose data with human-annotated sentences.
    Experimental results show that our model advances the state-of-the-art on text-based motion synthesis in objective evaluations by a margin of 50\%. Qualitative evaluations based on a user study indicate that our synthesized motions are perceived to be the closest to the ground-truth motion captures for both short and compositional sentences.
\end{abstract}

\begin{figure}[t]
\centering
    \includegraphics[width=\columnwidth]{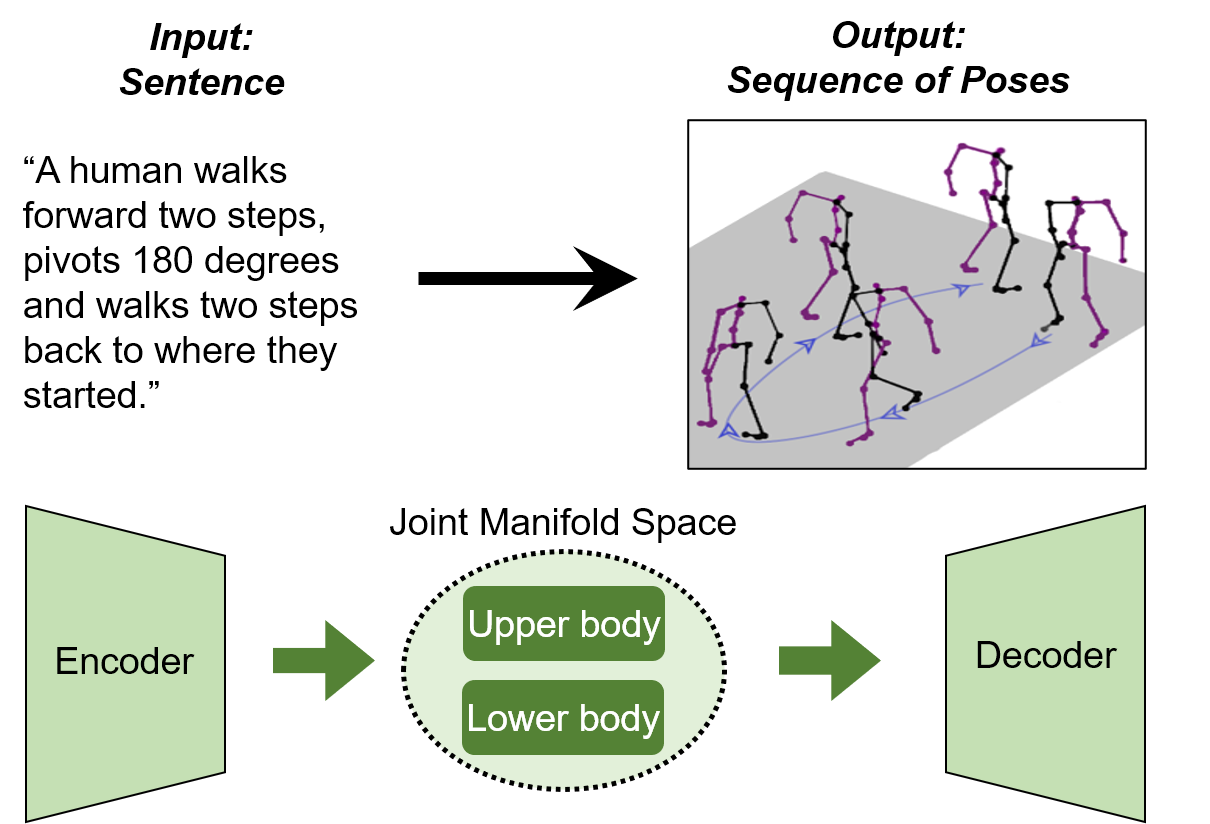}
    \caption{Overview of our proposed method to generate motion from complex natural language sentences.}
\label{fig:cover_img}
\vspace{-15pt}
\end{figure}
\section{Introduction}
Manually creating realistic animations of humans performing complex motions is challenging. Motion synthesis based on textual descriptions substantially simplifies this task and has a wide range of applications, including language-based task planning for robotics and virtual assistants~\cite{ahuja2020style}, designing instructional videos, creating public safety demonstrations~\cite{lafortune2018learning}, and visualizing movie scripts~\cite{hanser2009scenemaker}.
However, mapping natural language text descriptions to 3D pose sequences for human motions is non-trivial. The input texts may describe single actions with sequential information (\textit{e.g.}, ``a person walks four steps forward''), or may not correspond to the discrete time steps of the pose sequences to be generated, in case of superimposed actions (\textit{e.g.}, ``a person is spinning around while walking''). This necessitates a machine-level understanding of the syntax and the semantics of the text descriptions to generate the desired motions~\cite{l2p}.

While translating a sentence to a pose sequence, we need to identify the different parts of speech in the given sentence and how they impact the output motion. A verb in the sentence describes the type of action, whereas an adverb may provide information on the direction, place, frequency, and other circumstances of the denoted action. These need to be mapped into the generated pose sequence in the correct order, laying out additional challenges for motion modeling systems.
Existing text-to-motion mapping methods can either generate sentences describing one action only~\cite{bidirectional_mapping} or produce incorrect results for descriptions of compositional actions~\cite{l2p}. They fail to translate long-range dependencies and correlations in complex sentences and do not generalize well to other types of motions outside of locomotion~\cite{l2p}. We propose a method to handle complex sentences, meaning sentences that describe a person performing multiple actions either sequentially or simultaneously. For example, the input sentence ``a person is stretching his arms, taking them down, walking forwards for four steps and raising them again'' describes multiple sequential actions such as raising the arms, taking down the arms, and walking, as well as the direction and number of steps for the action. 
To the best of our knowledge, our method is the first to synthesize plausible motions from such varieties of complex textual descriptions, which is an essential next step to improve the practical applicability of text-based motion synthesis systems. To achieve this goal, we propose a hierarchical, two-stream, sequential network that synthesizes 3D pose sequences of human motions by parsing the long-range dependencies of complex sentences, preserving the essential details of the described motions in the process. Our output is a sequence of 3D poses generating the animation described in the sentence (Fig.~\ref{fig:cover_img}). Our main contributions in this paper are as follows:

\textbf{Hierarchical joint embedding space.}
In contrast to~\cite{l2p}, we separate our intermediate pose embeddings into two embeddings, one each for the upper body and the lower body. We further separate these embeddings hierarchically to limb embeddings. Our model learns the semantic variations in a sentence ascribing speed, direction, frequency of motion, and maps them to temporal pose sequences by decoding the combined embeddings. This results in the synthesis of pose sequences that correlate strongly with the descriptions given in the input sentences. 

\textbf{Sequential two-stream network.} 
We introduce a sequential two-stream network with an autoencoder architecture, with different layers focusing on different parts of the body, and combine them hierarchically to two representations for the pose in the manifold space -- one for the upper body and the other for the lower body. This reduces the smoothing of upper body movements (such as wrist movements for playing violin) in the generated poses and makes the synthesized motion more robust.

\textbf{Contextualized BERT embeddings.}
In contrast to previous approaches~\cite{l2p, bidirectional_mapping}, which do not use any contextualized language model, we use the state-of-the-art BERT model~\cite{devlin2018bert} with handpicked word feature embeddings to improve text understanding.

\textbf{Additional loss terms and pose discriminator.}
We add a set of loss terms to the network training to better condition the learning of the velocity and the motion manifold~\cite{motion_manifold}. We also add a pose discriminator with an adversarial loss to further improve the plausibility of the synthesized motions.

Experimental results show that our method outperforms the state-of-the-art methods of Ahuja et al.~\cite{l2p} and Lin et al.~\cite{lin2018generating} significantly on both the quantitative metrics we discuss in Section~\ref{subsection:quantitative_metrics} and on qualitative evaluations.

\section{Related Work}
This section briefly summarizes prior works in the related areas of data-driven human motion modeling and text-based motion synthesis.
\subsection{Human Motion modeling}
Data-driven motion synthesis  is widely used to generate realistic human motion for digital human models~\cite{holden2017phase,henter2020moglow,du2019stylistic}.
Different strategies have been implemented over the years using temporal convolutional neural networks~\cite{cui2021efficient,lea2017temporal,cheema2018dilated}, graph convolution networks~\cite{aksan2019structured,mohamed2020social} and recurrent neural networks~\cite{martinez2017human,habibie2017recurrent,wang2019combining,kundu2020cross}. Pose forecasting attempts to generate short~\cite{fragkiadaki2015recurrent, pavllo2018quaternet} and long-term motions~\cite{ghosh2017learning, li2017auto, tang2018long} by predicting future sequence of poses given their history.
Prior works encode the observed information of poses to latent variables and perform predictions based on the latent variables~\cite{motion_manifold,holden2015learning}. Holden et al.~\cite{holden2016deep} used a feed-forward network to map high-level parameters to character movement. Xu et al.~\cite{xu2020hierarchical} proposed a hierarchical style transfer-based motion generation, where they explored a self-supervised learning method to decompose a long-range generation task hierarchically.
Aristidou et al.~\cite{aristidou2018deep} break the whole motion sequences into short-term movements defining motion words and cluster them to a high-dimensional feature space. 
Generative adversarial networks~\cite{goodfellow2014generative} have also gained considerable attention in the field of unsupervised learning-based motion prediction~\cite{barsoum2018hp,kundu2019bihmp}.  Li et al.~\cite{li2018convolutional}
used a convolutional discriminator to model human motion sequences to predict realistic poses. Gui
et al.~\cite{gui2018adversarial} presents the adversarial geometry aware encoder-decoder (AGED) framework, where two global recurrent discriminators distinguish the predicted pose from the ground-truth.
Cui et al.~\cite{cui2020learning} propose a generative model for pose modeling based on graph networks and adversarial learning.

Related work also include pixel-level prediction using human pose as an intermediate variable~\cite{villegas2017learning, walker2017pose}, locomotion trajectories forecasting~\cite{hasan2018seeing,hasan2019forecasting,mangalam2020disentangling}. Various audio, speech, and image conditioned forecasting\cite{baltruvsaitis2018multimodal}  have also been explored for predicting poses. For instance,~\cite{ferreira2020learning} explores generating skeleton pose sequences for dance movements from audio,~\cite{chao2017forecasting, wu2019real} aims at predicting future pose sequences from static images.~\cite{ahuja2019coalescing} has linked pose prediction with speech and audio. Takeuchi et al.\cite{takeuchi2017speech} tackled speech conditioned forecasting for only the upper body, modeling the non-verbal
behaviors such as head nods, pose switches, hand waving for a character without providing knowledge on the character's next movements. ~\cite{chiu2019action} rely solely on the history of poses to predict what kind of motion will follow.

\begin{figure*}[t]
\centering
    \includegraphics[width=\textwidth]{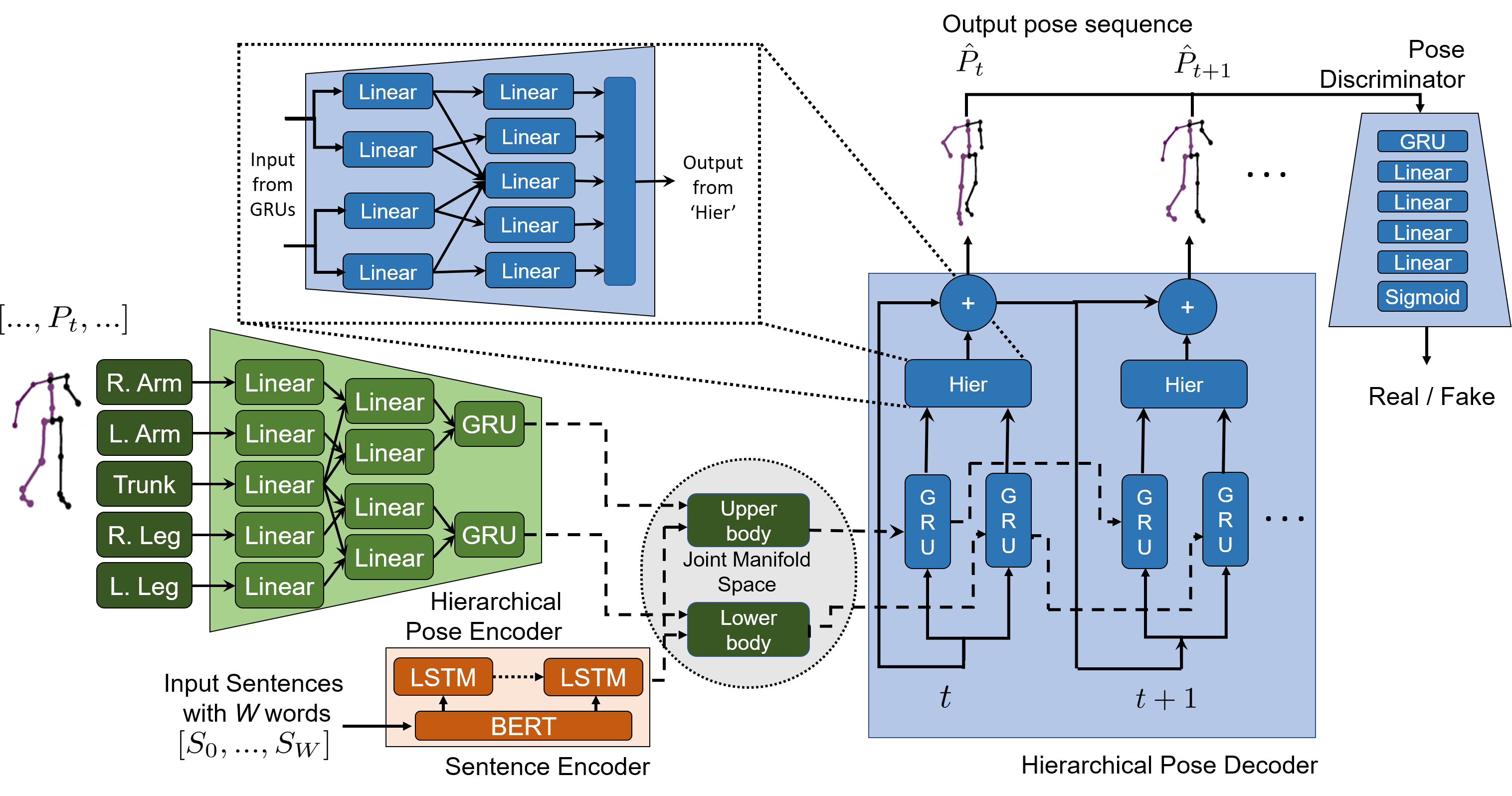}
    \caption{Structure of our hierarchical two-stream model along with pose discriminator. The model learns a joint embedding for both pose and language. The embedding has separate representations for the upper body and lower body movements.}
\label{fig:joint_model_fig}
\vspace{-15pt}
\end{figure*} 

\subsection{Text-based Motion Synthesis}
A subset of prior works have opted to train deep learning models to
translate linguistic instructions to actions for virtual agents~\cite{hatori2018interactively, hermann2017grounded, mei2016listen, yamada2016dynamical}.  Takano et al. describe a system that learns a mapping between human motion and word labels using Hidden Markov Models in~\cite{takano2015symbolically, takano2007interactive}. They also used statistical methods~\cite{takano2012bigram, takano2015statistical} using bigram models for natural languages to generate motions. Yamada et al.~\cite{yamada2018paired} use separate autoencoders for text and animations with a shared latent space to generate animations from text. 
Ahn et al.~\cite{text2action} generates actions from natural language descriptions for video data. However, their method only applies to upper-body joints (neck, shoulders, elbows, and wrist joints) with a static root. 
Recent methods mentioned in~\cite{bidirectional_mapping,lin2018generating,l2p} used RNN based sequential networks to map text inputs to motion.
Plappert et al.~\cite{bidirectional_mapping} propose a bidirectional RNN network to map text input to a series of Gaussian distributions representing the joint angles of the skeleton. However, their input sequence is encoded into a single one-hot vector that cannot scale as the input sequence becomes longer. Lin et al.~\cite{lin2018generating} use an autoencoder architecture to train on mocap data without language descriptions first, and then use an RNN to map descriptions into these motion representations.
Ahuja et al.~\cite{l2p} learns a joint embedding space for both pose and language using a curriculum learning approach. Training a model jointly with both pose and sentence inputs improves the generative power of the model. 
However, these methods are limited to synthesize motion from simple sentences. Our model, by contrast, handles long sentences describing multiple actions.

\section{Proposed Method}
We train our model end-to-end with a hierarchical two-stream pose autoencoder, a sentence encoder, and pose discriminator as shown in Fig.~\ref{fig:joint_model_fig}. Our model learns a joint embedding between the natural language and the poses of the upper body and the lower body.
Our input motion $P=[P_0,...,P_{T-1}]$ is a sequence of $T$ poses, where $P_t \in \mathbb{R} ^{J\times3 } $ is the pose at $t^{th}$ time step. $J \times 3$ indicate the joints of the skeleton with the $\parens{x, y, z}$ coordinates of each joint. Our hierarchical two-stream pose encoder $(pe)$ encodes the ground truth pose sequence $P$ into two manifold vectors,
\begin{equation}
    pe\parens{P} = (Z^p_{ub}, Z^p_{lb})
    \label{eq:pose_enc}
\end{equation}
where $Z^p_{ub}, Z^p_{lb} \in \mathbb{R}^h$ represent the features for the upper body and the lower body, respectively, and $h$ denotes the dimension of the latent space.

Our input sentence $S = [S_1, S_2,...,S_W]$ is a sequence of $W$ words converted to word embeddings $\tilde{S}_w$ using the pre-trained BERT model~\cite{devlin2018bert}. $\tilde{S}_w\in \mathbb{R}^K$ represents the word embedding vector of the $w^{th}$ word in the sentence and $K$ is the dimension of the word embedding vector used. Our two-stream sentence encoder $(se)$ encodes the word embeddings and maps them to the latent space such that we have two latent vectors, 
\begin{equation}
    se\parens{S} = (Z^s_{ub}, Z^s_{lb})
    \label{eq:sentence_encoding}
\end{equation}
where $Z^s_{ub}, Z^s_{lb} \in \mathbb{R}^h$ represent the sentence embeddings for the upper body and the lower body, respectively. Using an appropriate loss (see Section~\ref{subsection:loss}), we ensure that $(Z^p_{ub}, Z^p_{lb})$ and  $(Z^s_{ub}, Z^s_{lb})$ lie close in the joint embedding space and carry similar information.

Our hierarchical two-stream pose decoder $(de)$ learns to generate poses from these two manifold vectors. As an initial input, the pose decoder uses the initial pose $P_t$ of time step $t=0$ to generate the pose $\hat P_t$, which it uses recursively as input to generate the next pose $\hat P_{t+1}$. $\hat P \in \mathbb{R}^{T\times J\times 3}$ denotes a generated pose sequence. The output of our decoder module is a sequence of $T$ poses $\hat{P^p} \in \mathbb{R} ^{T\times J\times 3}$ generated from the pose embeddings, and $\hat{P^s} \in \mathbb{R} ^{T\times J\times 3 }$ generated from the language embeddings:
\begin{align}
    \hat{P^p} &= de\parens{Z^p_{ub}, Z^p_{lb}} \\
    \hat{P^s} &= de\parens{Z^s_{ub}, Z^s_{lb}}.
    \label{eq:decode_from_sentence}
\end{align}
We use a pose prediction loss term to ensure that $\hat{P^p}$ and $\hat{P^s}$ are similar (Section~\ref{subsection:loss}). 
$\hat P = \hat{P^s}$ is our final output pose sequence for a given sentence.

\subsection{Network Architecture}
The three main modules in our network are the two-stream hierarchical pose encoder, the two-stream sentence encoder and the two-stream hierarchical pose decoder. We explain the architecture of all these modules.

\subsubsection{Two-Stream Hierarchical Pose Encoder}
We structure the pose encoder such that it learns features from the different components of the body. Individual parts are then combined hierarchically. We decompose the human skeleton into the five major parts - left arm, right arm, trunk, left leg, and right leg as done in~\cite{du2015hierarchical}.
Our hierarchical pose encoder, as shown in Fig~\ref{fig:joint_model_fig}, encodes these five parts using five linear layers with output dimension $h_1$. We combine the trunk representation with that of the left arm, right arm, left leg, and right leg and pass them through another set of linear layers to obtain combined representations of (left arm, trunk), (right arm, trunk), (left leg, trunk), and (right leg, trunk) each of dimension $h_2$ each. Two separate GRUs~\cite{cho2014properties} encode the combined representation for the arms with trunk and the legs with trunk respectively, thus creating two manifold representations -- one for the upper body $(Z^p_{ub} \in \mathbb{R}^h)$ and the other for the lower body $(Z^p_{lb} \in \mathbb{R}^h)$.
The output of the GRUs give the two manifold representations of dimension $h$.
\subsubsection{Two-Stream Sentence Encoder}
\label{subsubssection:SE}
To represent the text input, we use the pre-trained large-case model of BERT~\cite{devlin2018bert} as a contextualized language model. It comprises 24 subsequent layers, each representing different linguistic notions of syntax or semantics~\cite{clark2019does}. To find the focused layers on local context (\textit{e.g.}, adverbs of a verb)~\cite{tenney2019you}, we use the attention visualization tool~\cite{vig2019multiscale} with randomly selected samples of the KIT Motion Language dataset~\cite{KitPlappert2016}.
Thus, we select the layers $12$ (corresponding to subject(s)), $13$ (adverb(s)), $14$ (verb(s)) and $15$ (prepositional object(s)) and concatenate the hidden states of these layers in order to represent the corresponding word.
Formally, $\tilde{S}_w\in \mathbb{R}^K$ represents the word embedding vector of the $w^{th}$ word in the sentence $S$, and $K$ is the dimension of the word embedding vector used.
Our Sentence encoder $(se)$ uses Long-Short Term Memory units (LSTMs)~\cite{schmidhuber1997long} to capture the long-range dependencies of a complex sentence. We input the word embeddings to a two-layer LSTM, which generates $Z^s \in \mathbb{R}^{2h}$, where, 
\begin{equation}
    LSTM\parens{\tilde{S}_w} = Z^s = [Z^s_{ub}, Z^s_{lb}]
\end{equation}
is the latent embedding of the whole sentence, with $\tilde{S}_w = BERT(S_w)$.
We use the first half of this embedding as $Z^s_{ub} \in \mathbb{R}^h$ to represent the upper body and the second half as $Z^s_{lb} \in \mathbb{R}^h$ to represent the lower body.

\subsubsection{Two-Stream Hierarchical Pose Decoder}
We can conceptually unfold our pose decoder as a series of $T$ hierarchical decoder units, each constructing the output pose $\hat{P}_t$, $\forall t = 0, \dots, T$ time steps in a recurrent fashion by taking in the generated pose at the corresponding previous time step. We add a residual connection between the input and the output of the individual decoder units as shown in Fig.~\ref{fig:joint_model_fig}. 
Each decoder unit consists of two GRUs, and a series of linear layers structured hierarchically. The hierarchical structure of the linear layers in the decoder unit mirrors that of the pose encoder. Conditioned by the latent space vector representing the previous frames, the GRUs and the hierarchical linear layers \textit{Hier} (as shown in Fig~\ref{fig:joint_model_fig}) output the reconstructed pose $\hat{P}_{t+1}$ at the $(t+1)^{th}$ frame given its previous pose $\hat{P}_t$. 

\subsection{Optimizing the Training Procedure}
\label{subsection:loss}
We train our model end-to-end with a hierarchical two-stream pose autoencoder along with a sentence encoder as shown in Fig.~\ref{fig:joint_model_fig}. Our model learns a joint embedding space between the natural language and the poses of the upper body and the lower body. Our decoder is trained twice in each pass: once with $(Z^p_{ub}, Z^p_{lb})$ obtained from $pe$ to generate the pose sequence $\hat P^p$, and the second time with the $(Z^s_{ub}, Z^s_{lb})$ obtained from $se$, which generates the pose sequence $\hat P = \hat P^s$.

\textbf{Loss functions.} We use the smooth $\ell_1$ loss as a distance metric to train our model. The smooth $\ell_1$ loss is less sensitive to outliers than the smoother $\ell_2$ loss, and more stable than the $\ell_1$ loss as it is differentiable near $x=0$ for all $x\in \mathbb{R}$~\cite{l2p}. We use the following losses while training the whole model: 
\begin{itemize}
    \item \textbf{Pose Prediction loss}: It minimizes the difference between the input ground-truth motion $(P)$ and the predicted motions $\hat P = \hat{P}^s$ and $\hat{P}^p$. We measure it as,
    \begin{equation}
        L_R = \mathcal{L} \parens{\hat{P}^s, P} + \mathcal{L} \parens{\hat{P}^p, P},
    \end{equation}
    where $\mathcal{L}$ denotes the Smooth $\ell_1$ Loss between the two terms.
    \item \textbf{Manifold reconstruction loss}: This encourages a reciprocal mapping between the generated motions and the manifold representations to improve the manifold space~\cite{motion_manifold}. For that, we reconstruct the manifold representations from the generated poses as $\hat{Z}^p_{ub} = pe\parens{\hat{P}}$ and $\hat{Z}^p_{lb} = pe\parens{\hat{P}}$, and compare them with the manifold representations obtained from input pose sequence. We compute the loss as,
    \begin{equation}
        L_M = \mathcal{L}\parens{\hat{Z}^p_{ub}, Z^p_{ub}} + \mathcal{L}\parens{\hat{Z}^p_{lb}, Z^p_{lb}}.
    \end{equation}
    \item \textbf{Velocity reconstruction loss}: We minimize the difference between the velocity of the reconstructed motion  $\parens{\hat{P}_{vel}}$ and the velocity of the input motion $\parens{P_{vel}}$. We compute the velocity of the $t^{th}$ frame of a pose $P$ as $P_{vel}(t) = P_{(t+1)} - P_{(t)}$. We compute $L_V$ as , 
    \begin{equation}
        L_V = \mathcal{L}\parens{\hat{P}_{vel}, P_{vel}}. 
    \end{equation}
    \item \textbf{Embedding similarity loss:} We use this loss to ensure that the manifold representations, $Z^s_{ub}$ and $Z^s_{lb}$, generated by the sentence encoder is similar to the manifold representations $Z^p_{ub}$ and $Z^p_{lb}$ generated by the pose encoder. We measure it as,
    \begin{equation}
        L_E = \mathcal{L}\parens{Z^p_{ub}, Z^s_{ub}} + \mathcal{L}\parens{Z^p_{lb}, Z^s_{lb}}.
        \label{eq:encoder_loss}
    \end{equation}
    \item \textbf{Adversarial loss:} We further employ a binary cross-entropy discriminator $D$ to distinguish between the real and generated poses. We compute the corresponding discriminator and ``generator" losses as,
    \begin{align}
        L_D &= \mathcal{L}_2\parens{D\parens{\hat{P}}, 0} + \mathcal{L}_2\parens{D\parens{P}, 1}  \\
        L_G &= \mathcal{L}_2\parens{D\parens{\hat{P}}, 1},
    \end{align}
    where $\mathcal{L}_2$ denotes the Binary Cross Entropy loss, and the ``generator" is the decoder of the auto-encoder.  
\end{itemize}
We train the model end-to-end with the pose autoencoder, the sentence encoder and the discriminator modules on a weighted sum of these loss terms as,
\begin{align}
    &\underset{pe, se, de}{\text{min}} \parens{L_R + \lambda_M L_M + \lambda_V L_v + \lambda_E L_E + \lambda_G L_G}& \nonumber \\
    &\underset{D}{\text{min}}\parens{\lambda_G L_D},&
\end{align}
where $\lambda_M=0.001$, $\lambda_V = 0.1$, $\lambda_E = 0.1$ and $\lambda_G=0.001$ are weight parameters, obtained experimentally.

\section{Experiments}
This section describes the dataset we use for our experiments and report the quantitative and qualitative performance of our method. We also highlight the benefits of the different components of our method via ablation studies.

\subsection{Dataset}
We evaluate our model on the publicly available KIT Motion-Language Dataset~\cite{KitPlappert2016} which consists of $3,911$ recordings of human
whole-body motion in MMM representation~\cite{MMMterlemez2014master, mandery2016unifying} with natural language descriptions corresponding to each motion. There is a total of $6,278$ annotations in natural language, with each motion recordings having one or multiple annotations describing the task. The sentences range from describing simple actions such as walking forwards or waving the hand to describing motions with complicated movements such as waltzing. Moreover, there are longer, more descriptive sentences describing a sequence of multiple actions, \textit{e.g.}, ``\textit{A human walks forwards two steps, pivots 180 degrees and walks two steps back to where they started}". 
We split the whole dataset into random samples in the ratio of $0.6$, $0.2$, and $0.2$ for training, validation, and test sets.
For better comparison with the state-of-the-art~\cite{l2p,lin2018generating}, we pre-process the given motion data in the same manner as done in~\cite{l2p,lin2018generating}. Following the method of Holden et al.~\cite{holden2016deep}, we use the character's joint positions with respect to the local coordinate frame and the character’s trajectory of movement in the global coordinate frame.
We have $J=21$ joints, each having $\parens{x, y, z}$ coordinates, and a separate dimension for representing the global trajectory for the root joint.
Similar to~\cite{l2p,lin2018generating}, we sub-sample the motion sequences to a frequency of $12.5$ Hz from $100$ Hz.

\begin{figure*}[t]
\centering
    \includegraphics[width=\textwidth]{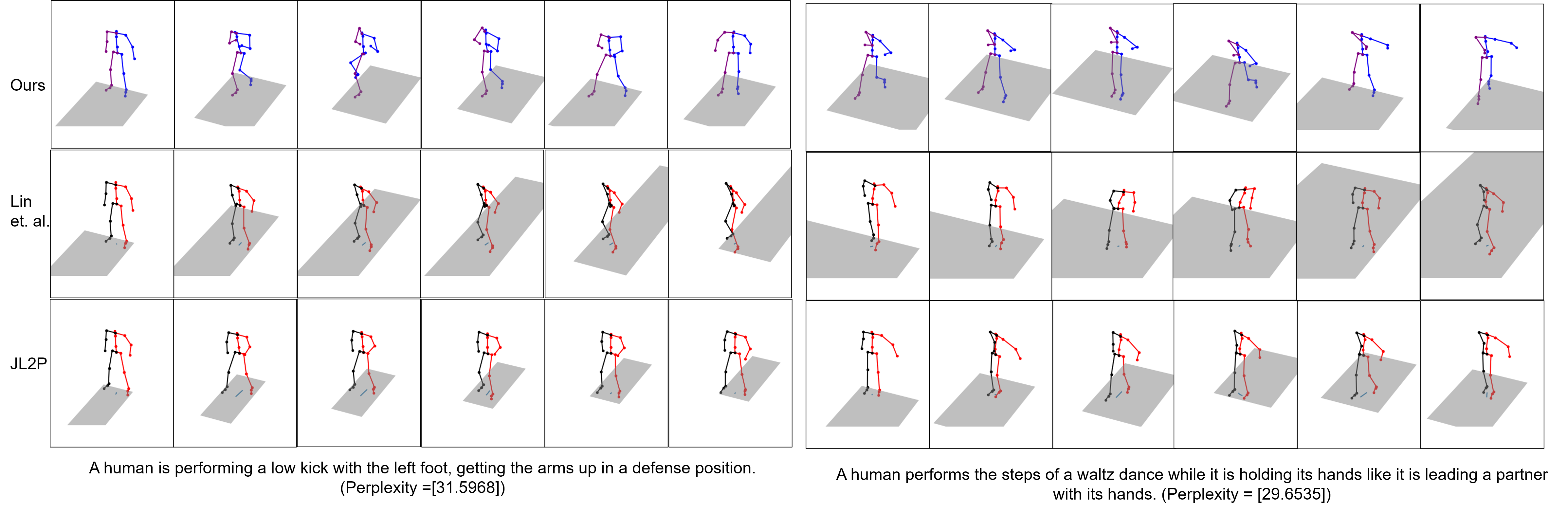}
\caption{Comparison of consecutive frames of generated animations of our method (top row) with Lin et al.~\cite{lin2018generating} (middle row) and JL2P~\cite{l2p} (bottom row) for the given sentences. Our method generates clear kicking and dancing motions in contrast to JL2P and Lin et al., which do not show any prominent movements. The perplexity values of the sentences are according to~\cite{KitPlappert2016}.}
\label{fig:plot_all}
\vspace{-15pt}
\end{figure*}

\subsection{Implementation Details}
We trained our model for $350$ epochs using Adam Optimizer.  Total training time was approximately $20$ hours using an NVIDIA Tesla V100. The dimensions of our hidden layers in the hierarchical autoencoder are $h_1=32 $, $h_2=128 $ and $h=512$.  We used a batch size of $32$ and a learning rate of $0.001$ with exponential decay. For training the sentence encoder, we converted given sentences to word embeddings of dimension $K=4,096$ using selected layers of the pre-trained BERT-large-case model (details in~\ref{subsubssection:SE}). We encoded these embeddings to a dimension of $1024$ through the sentence encoder, and split it to obtain two manifold representations $Z^s_{ub} \in \mathbb{R}^h$ and $Z^s_{lb} \in \mathbb{R}^h$, each of dimension $h=512$. 

\subsection{Quantitative Evaluation Metrics}
\label{subsection:quantitative_metrics}
To quantitatively evaluate the correctness of our motion, we use the Average Position Error (APE). APE measures the average positional difference for a joint $j$ between the generated poses and the ground-truth pose sequence as,
\begin{equation}
    APE[j] = \frac{1}{NT} \sum_{n\in N} \sum_{t\in T} \norm{P_t[j] - \hat{P}_t[j]}_2,
\end{equation}
where $T$ is the total time steps and $N$ is the total number of data in our test dataset.

Given our setting of natural language descriptions and corresponding free-form movements, it is naturally difficult to find a quantitative measure that does justice to both modalities. For example, in a walking setting, sentences that do not mention any direction correspond to a wider variety of plausible motions, while specifying a direction narrows the possibilities. To account for such discrepancies, we separate the APEs between the local joint positions and the global root trajectory. The former corresponds to the error of the overall poses, while the latter corresponds to the overall direction and trajectory of the motion.

However, the average position of each joint simply corresponds to a mean compared to the dataset. To understand the full statistics of the overall distribution compared to the dataset, we also compute the Average Variance Error (AVE), which measures the difference of variances of individual joints of the generated poses compared to the ground truth poses. We calculate the variance of an individual joint $j$ for a pose $P$ with $T$ time steps as,
\begin{equation}
    \sigma[j] = \frac{1}{T-1}\sum_{t \in T}\parens{P_t[j] - \tilde{P}[j]}^2,
\end{equation}
where $\tilde{P}[j]$ is the mean pose over $T$ time steps for the joint $j$.
Calculating the variance for all joints of the ground-truth poses and the generated poses, we use their root mean square error as the AVE metric as follows:
\begin{equation}
    AVE[j] = \frac{1}{N}\sum_{n\in N} \norm{\sigma[j] - \hat\sigma[j]}_2,
\end{equation}
where $\sigma$ refers to the ground-truth pose variance and $\hat\sigma$ refers to generated pose variance.

However, even this measure does not account for any information regarding the sentences or sentence encodings themselves. Therefore, we propose a Content Encoding Error (CEE), which corresponds to the embedding similarity loss $L_E$ in Eq.~\ref{eq:encoder_loss} by measuring the effectiveness of the embedding space. We calculate CEE as the difference between manifold representations $Z^p=[Z^p_{ub},Z^p_{lb}]$ (obtained by encoding the input poses $P$ through the pose encoder $pe$) and the manifold representations $Z^s=[Z^s_{ub},Z^s_{lb}]$ (obtained by encoding the corresponding input sentences using the sentence encoder $se$). We write it as,
\begin{equation}
    CEE(S, P) = \frac{1}{MN} \sum_{n\in N} \sum_{m \in M}\norm{Z^s - Z^p}_2,
\end{equation}
where $M$ is the number of features in the manifold representation, and $N$ is the total number of data. The idea is to measure how well the joint embedding space correlates the latent embeddings of poses with the latent embeddings of the corresponding sentences.

To further account for style factors in the motion and the sentences, we propose a Style Encoding Error (SEE). SEE compares a summary statistics of the sentence embeddings $Z^s$ and the pose embeddings $Z^p$ to account for general style information. We compute the Gram matrix~\cite{gatys2016image, gatys2015texture} $G$ on the corresponding embeddings:

\begin{align}
    G_s &= Z^s \cdot Z^{s^\top}\\
    G_p &= Z^p \cdot Z^{p^\top}
\end{align}

We compute SEE as:

\begin{equation}
    SEE(S, P) = \frac{1}{MN} \sum_{n\in N} \sum_{m\in M}\norm{G_s - G_p}_2,
\end{equation}
where $M$ is the number of features in the manifold representation and $N$ is the total number of data.

\begin{figure*}[ht]
\centering
    \includegraphics[width=\textwidth] {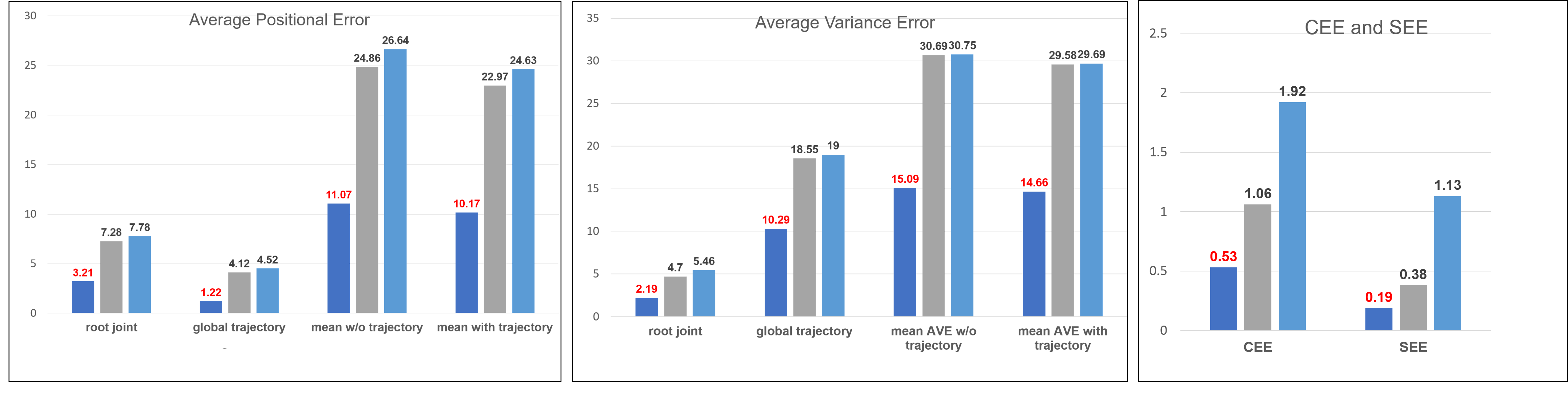}
\caption{Plots showing the APE (left), AVE (middle), CEE and SEE (right) in mm for our model compared to the JL2P~\cite{l2p}, Lin et al.~\cite{lin2018generating}. Dark blue line denotes our method, grey denotes JL2P and light blue denotes Lin et al. method. Lower values are better. We see our method improves over the state-of-the-art by over $50$\% on all benchmarks.}
\label{fig:quant_all}
\vspace{-15pt}
\end{figure*}

\subsection{Ablation Studies}
\label{sec:ablations}
We compare the performance of our model with the following four ablations acquired from itself:
\begin{itemize}
    \item \textbf{Ablation 1: Two-stream hierarchical model without jointly training the embedding space (w/o JT).} Instead of end-to-end training of the model, we trained the hierarchical pose encoder and decoder first, using the loss terms $L_R$, $L_M$, $L_V$, $L_G$ and $L_D$ (discussed in Section~\ref{subsection:loss}). We then trained the model with the sentence encoder and the pose decoder with $L_R$ and $L_E$. This indicates that the model is not learning a joint embedding space for pose and language but learns the embedding space for poses first and then fine-tunes to map the sentences. 
    \item \textbf{Ablation 2: Hierarchical model without the two-stream representation (w/o 2-St).} We used a single manifold representation for the whole body instead of separating the upper and lower body and trained the model jointly on language and pose inputs.
    \item \textbf{Ablation 3: Training the hierarchical two-stream model without the extra losses (w/o Lo).} 
    We discarded the additional loss terms introduced in the paper in Section~\ref{subsection:loss} and only used the pose prediction loss $L_R$ to train our model. 
    \item \textbf{Ablation 4: Using a pre-trained language model instead of selected layers of BERT (w/o BERT).}
    We used a pre-trained Word2Vec model~\cite{mikolov2013distributed} as done in~\cite{l2p} to convert the input sentence into word embeddings instead of selecting layers of BERT as mentioned in Section~\ref{subsubssection:SE}. 
    This ablation shows how BERT as a contextualized language model, helped to focus on the local context within a sentence.  
\end{itemize}

\subsection{User Study}
\begin{figure}[t]
\centering
    \includegraphics[width=\columnwidth]{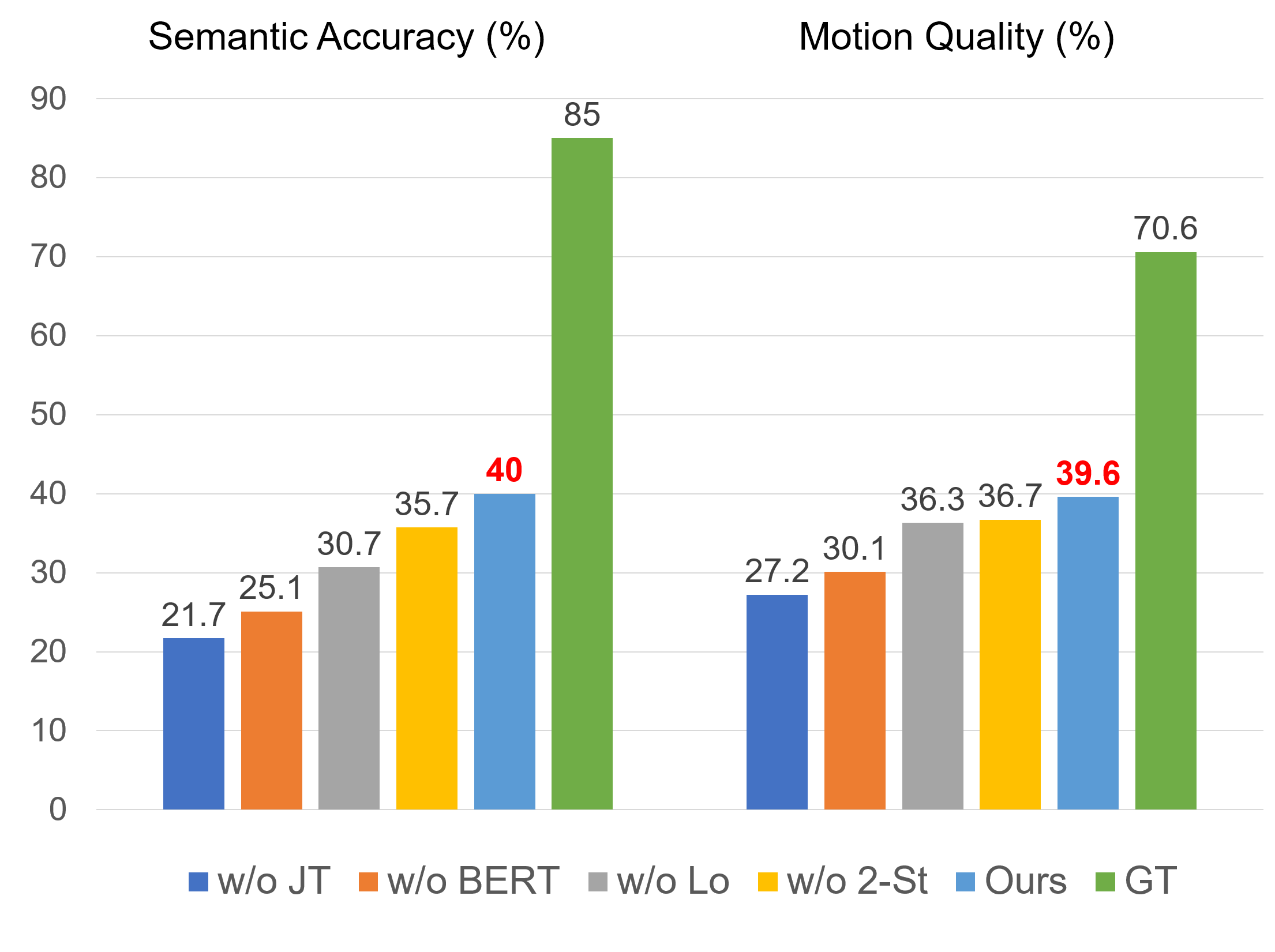}
\caption{Semantic accuracy in percentage denoting how good the motion visually corresponds to the input sentences (left) and Motion quality in percentage showing how good the overall quality of motion is in terms of naturalness (right). Higher value is better. Our method's score denoted in red, has highest percentage compared to the ablations.}
\label{fig:userstudy}
\vspace{-15pt}
\end{figure}

To evaluate our ablation studies, we conducted a user study to observe the subjective judgment of the quality of our generated motions compared to the quality of motions generated from the ablations described in Section~\ref{sec:ablations}. We asked 23 participants to rank 14 motion videos from the five methods and from the ground-truth motion-captures, based on whether the motion corresponds to the input text, and by the quality and naturalness of the motions. The five methods include our method and the four ablations of our model -- `w/o JT', `w/o 2-St', `w/o Lo', and `w/o BERT'. 
We quantified the user study with two preference scores -- the first one describing if the participants found the motions to correspond to the input sentence (``yes/no"), and the second one rating the overall quality of the motion in terms of naturalness (from $1=$``most natural" to $6=$``least natural", which we then scaled to 0 and 1 and inverted). We observe that our method has a preference score of $\sim40$\% in both cases, second only to the ground truth motion as seen in Fig.~\ref{fig:userstudy}. 
~\footnote{We decided to exclude~\cite{l2p} and~\cite{lin2018generating} from the user study, based on overwhelming feedback from participants that our method beats the state-of-the-art in the most obvious ways. We provide additional qualitative results for these in the supplementary material.}

\section{Results and Discussion}
\label{sec:results}
We compare our method with the state-of-the-art Joint Language to Pose (JL2P) method~\cite{l2p}, and the proposed approach by Lin et al.~\cite{lin2018generating}. We have used the pre-trained models for both JL2P and Lin et al.'s approach, provided by Ahuja et al.~\cite{l2p}, to calculate the quantitative results. We computed all the results on the test dataset.

\subsection{Objective Evaluation}
Fig.~\ref{fig:quant_all} shows the improvement of our method compared to JL2P and Lin et al. for all the metrics discussed in Section~\ref{subsection:quantitative_metrics}. Our method shows an improvement of $55.4$\% in the mean APE calculated for all local joints compared to JL2P and by $58.4$\% compared to Lin et al. When included with the global trajectory, our method still shows an improvement of $55.7$\% in mean APE compared to JL2P and an improvement of $58.7$\% in mean APE compared to Lin et al. (Fig.~\ref{fig:quant_all} left).~\footnote{We note that our reported numbers for the state-of-the-art methods in the APE metric are different from the original paper. However, we were unable to replicate the numbers in the original paper using the code and the pre-trained model provided by the authors.} 

We also observe that high error in the root joint leads to either foot sliding in the motion or averages out the whole motion. Improvement in the error values for the root joint indicates high-quality animations without any artifacts like foot sliding.
Furthermore, our method shows closer resemblances to the variance of the ground truth motion compared to the state-of-the-art models (Fig.~\ref{fig:quant_all} center). Our method has an improvement of $50.4$\% in the AVE over the mean of all joints with global trajectory compared to JL2P, and an improvement of $50.6$\% over the mean of all joints with global trajectory compared to Lin et al. 
We provide detailed APE and AVE values of individual joints in the supplementary material.

We also show improvements of $50$\% in the CEE and SEE metrics compared to JL2P. Compared to Lin et al., we show improvements of $72.3$\% and $83.1$\% in the CEE and SEE, respectively (Fig.~\ref{fig:quant_all} right). These results show that the joint embedding space learned by our method can correlate the poses and corresponding sentences better than the state-of-the-art methods. 

\subsection{Qualitative Results}
\begin{figure}[t]
\centering
    \includegraphics[width=\columnwidth]{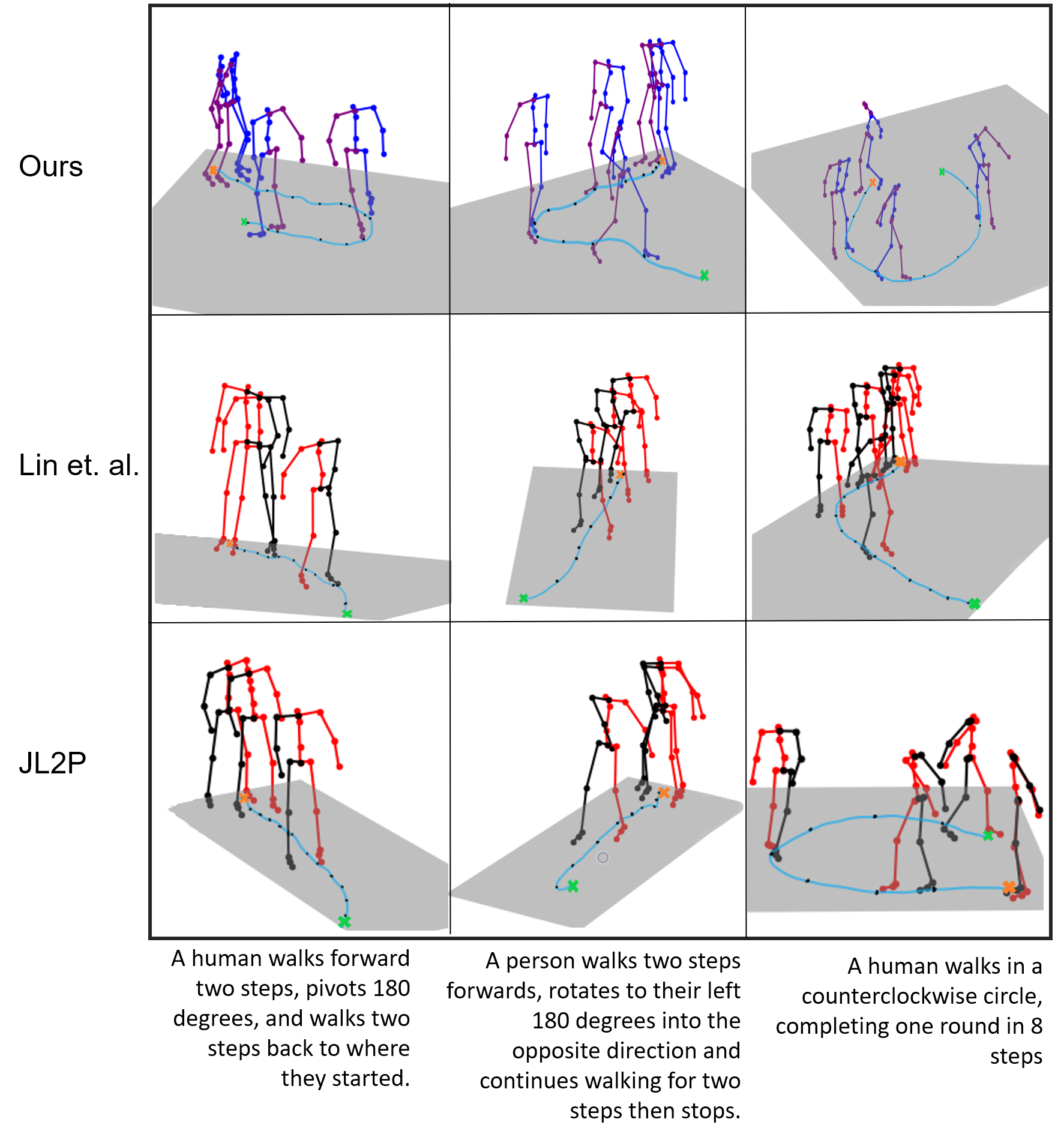}
\caption{Comparison of generated animations of our method (top row) with Lin et al.~\cite{lin2018generating} (middle row) and Ahuja et al.~\cite{l2p} (bottom row) for long sentences indicating direction and number of steps. Orange cross denotes starting point and green denotes end point of the motion. Blue line on the plane is the trajectory and the black dots represent the foot steps. Our method is clearly able to follow the semantics of the sentences, while the state-of-the-art fail.}
\label{fig:plot_trajectory}
\vspace{-15pt}
\end{figure}

To qualitatively compare our best model against the state-of-the-art methods~\cite{l2p, lin2018generating}, we examine the generated motions from all the methods. Fig.~\ref{fig:plot_all} shows two motions with rather high perplexity~\cite{KitPlappert2016} compared to the average movements in the dataset. Our method (top row) accurately generates the kicking action with the correct foot and right arm positions as described in the sentence, while the benchmark models fail to generate a kick at all (left). Fig.~\ref{fig:plot_all} (right) further shows that the Waltz dance is more prominent in our model, compared to both benchmarks where arm movements seem to be missing completely, and the skeleton tends to slide than actually step. Fig.~\ref{fig:plot_trajectory} shows screenshots with motions generated from rather complex sentence semantics. Our method (top row) accurately synthesizes a trajectory that matches the semantics of the sentence. Although Ahuja et al.~\cite{l2p} generate a circular trajectory (bottom right), their walking direction does not match the semantics of the sentence, while Lin et al.~\cite{lin2018generating} fail to generate a circular trajectory at all. Both methods also cannot synthesize correct turning motions (Fig.~\ref{fig:plot_trajectory} left and center columns).

\section{Limitations, Future Work and Conclusion}
We presented a novel framework that advances the state-of-the-art methods on text-based motion synthesis on qualitative evaluations and several objective benchmarks. While our model accurately synthesizes superimposed actions it encountered during training, generalization to novel superimpositions is not always successful, however. We intend to extend our model into a zero- or few-shot paradigm~\cite{socher2013zero} such that it generates simultaneous actions from input sentences without being trained on those specific combinations. We also plan to experiment with narration-based transcripts that describe long sequences of step-by-step actions involving multiple people, \textit{e.g.}, narration-based paragraphs depicting step-by-step movements for performing complex actions such as dance, work-outs, or professional training videos. To this end, a different embedding that explicitly models the sequential nature of the task may be more suitable. However, that may reduce the model's ability to synthesize actions not described in an exact sequential manner. Furthermore, improvements on general motion quality, such as foot sliding, limb constraints, and biomechanical plausibility, can be improved by introducing physical constraints~\cite{shimada2020physcap} to the model. 

Being able to model a variety of motions and handle such complex sentence structures is an essential next step in generating realistic animations for mixtures of actions in the long-term and improving the practical applicability of text-based motion synthesis systems. To the best of our knowledge, this is the first work to achieve this quality of motion synthesis on the benchmark dataset and is an integral step towards script-based animations.

\small{\paragraph{Acknowledgements.} We would like to thank all the participants in our user study, as well as the XAINES partners for their valuable feedback. This research was funded by the BMBF grants XAINES (01$|$W20005) and IMPRESS (01$|$S20076), as well as the EU Horizon 2020 grant Carousel+ (101017779) and an IMPRS-CS Fellowship. Computational resources were provided by the BMWi under the grants 01MK20004D and 01MD19001B.}

{\small
\bibliographystyle{ieee_fullname}
\bibliography{arxiv}
}
\begin{table*}[t]
    \centering
    \caption{Average Positional Error (APE) in mm for our model compared to the JL2P~\cite{l2p}, Lin et al.~\cite{lin2018generating}, and four ablations of our method described in Section 4.4 of our paper. Although the over all APE is lower for our ablation studies, we find the overall motion quality to be poorer than our final method due to larger errors in the root. Please refer to Section 5.1 of our paper for details.}
    \label{Tab:APE}
    \begin{tabular}{lrrrrrrr}
    \toprule
     & JL2P & Lin et al.& w/o BERT & w/o JT & w/o 2-St & w/o Lo &  Ours \\
    \midrule
    Trajectory & 4.12 & 4.52 & \textbf{1.21} & 1.27 & 1.22 & 1.23 & 1.22\\
    Root  &  7.28 & 7.78 & 3.23 & 3.50 & 3.22 & 3.23 &\textbf{3.21}\\
    Torso &  13.18 & 14.93 & 5.84 & \textbf{5.71} & \textbf{5.71} & 5.91 & 5.90\\
    Pelvis & 14.92 & 16.10 & \textbf{6.49} & 6.54 & 6.52 & 6.67 & 6.60 \\
    Neck & 33.01 & 36.03 & 14.88 & \textbf{14.50} & 14.69 & 15.04 & 15.01\\
    Left Arm & 37.37 & 41.71 & 16.54 & 16.79 & \textbf{16.09} & 16.79  & 16.94 \\
    Right Arm  &  37.91 & 42.33 & 16.41 & 16.56 & \textbf{15.81}  & 16.25 & 16.40 \\
    Left Hip &  13.50 & 14.33 & \textbf{6.02} & 6.12 & 6.14 & 6.18 & 6.21 \\
    Right Hip & 13.39 & 14.05 & \textbf{6.00} & 6.15 & 6.15 & 6.20 & 6.22 \\
    Left Foot &  38.38 & 38.84 & 16.78 & 16.63 & 16.84 & \textbf{16.25}  & 16.97 \\
    Right Foot & 39.66 & 40.31 & 17.12 & 17.15 & 17.24 & \textbf{16.78} & 17.22 \\
    \midrule
    Mean w/o trajectory & 24.86 & 26.64 & 10.93 & 10.96 & \textbf{10.84}  &
    10.93 & 11.07 \\
    \midrule
    Mean & 22.97 & 24.63 & 10.04 & 10.08 & \textbf{9.97} & 10.05 & 10.17 \\
    \bottomrule
    \end{tabular}
\end{table*}
\begin{table*}[t]
    \centering
    \caption{Average Variance Error (AVE) for our model compared to the JL2P~\cite{l2p}, Lin et al.~\cite{lin2018generating}, and the four ablations of our method described in Section 4.4 of our paper. Our method has the lowest AVE for the root joint as well as the mean of all the joints with and without the global trajectory.}
    \label{Tab:AVE}
    \begin{tabular}{lrrrrrrr}
    \toprule
     & JL2P & Lin et al. & w/o BERT & w/o JT & w/o 2-St & w/o Lo & Ours \\
    \midrule
    Trajectory & 18.55 & 19.00 & 10.87 & 10.52 & 11.20 & \textbf{9.75} & 10.29\\
    Root & 4.70 & 5.46 & 2.45 & 2.42 & 2.32 & 2.30 &  \textbf{2.19}\\
    Torso & 21.44 & 22.61 & 12.65 & 12.20 & 13.22 &  \textbf{11.85} & 11.87\\
    Pelvis & 23.79 & 24.51 & 13.66 & 13.25 & 13.99 & 12.73 &  \textbf{12.58}\\
    Neck & 45.05 & 36.03 & 26.24 & 25.26 & 27.37 & 24.78 & \textbf{24.65}\\
    Left Arm & 32.66 & 41.71 & 16.59 & 16.42 & 16.86 & 15.66 & \textbf{15.20}\\
    Right Arm & 29.15 & 42.34 & 15.18 & 14.54 & 15.05 & 14.31 & \textbf{13.95}\\
    Left Hip & 27.79 & 28.73 & 16.01 & 15.45 & 15.82 &  \textbf{14.35} & 14.71\\
    Right Hip & 26.73 & 27.05 & 14.46 & 14.13 & 14.92 &  \textbf{13.31} & 13.40\\
    Left Foot & 48.34 & 38.84 & 24.63 & 24.03 & 23.67 & 22.27 & \textbf{21.57}\\
    Right Foot & 47.23 & 40.31 & 23.04 & 23.10 & 22.80 & \textbf{20.72} & 20.87\\
    \midrule
    Mean w/o Trajectory & 30.69 & 30.75 & 16.49 & 16.08 & 16.60 & 15.22 & \textbf{15.09} \\
     \midrule
    Mean  & 29.58 & 29.69 & 15.98 & 15.57 & 16.11 & 14.73 & \textbf{14.66}\\
    \bottomrule
    \end{tabular}
\end{table*}

\balance
\section*{Appendix: More Results on Quantitative Evaluation Metrics}
We show the average positional error (APE) values for individual joints in Table~\ref{Tab:APE}. We compare our method with the two state-of-the-art methods~\cite{l2p, lin2018generating} and also with the four ablations of our method: `w/o BERT', `w/o JT', `w/o 2-St', `w/o Lo', as described in Section $4.4$ of our paper. We observe that high error in the root joint leads to either foot sliding in the motion or averages out the whole motion. Improvement in the error values for the root joint indicates high-quality animations without any such artifacts. When compared to the ablations of our model, we find that the APE calculated over the mean of all the joints with the global trajectory is marginally better for the ablations compared to our method (best for the ablation `w/o 2-St' showing an improvement of $1.96$\% over our method). 
This is because the motions get averaged out in the ablations, bringing the joint positions closer to the mean but reducing the relevant joint movements.
However, our method has the lowest APE for the root joint, implying that the overall motion quality is better. 
Using the additional metric of the average variance error (AVE) for evaluating the variability of the motions further shows that the joint movements are reduced in the ablations. Our method has the lowest AVE for the root joint as well as the mean of all the joints with and without the global trajectory, as shown in Table~\ref{Tab:AVE}. 
Our method also performs the best in terms of the content encoding error (CEE) and the style encoding error (SEE) compared to the ablations and the state-of-the-art methods as seen in Table~\ref{Tab:CEE}.

\begin{table*}[t]
    \centering
    \caption{Content Encoding Error (CEE) and Style Encoding Error (SEE) for our model compared to the JL2P~\cite{l2p}, Lin et al.~\cite{lin2018generating}, and the four ablations of our method described in Section 4.4 of our paper. Lower values are better. Our method has the lowest CEE and SEE.}
    \label{Tab:CEE}
    \begin{tabular}{lrrrrrrr}
    \toprule
    Method & JL2P & Lin et al. & w/o BERT & w/o JT & w/o 2-St & w/o Lo & Ours \\
    \midrule
    CEE & 1.06 & 1.92 & 1.10 & 0.99 & 0.67 & 1.04 & \textbf{0.53} \\
    SEE & 0.38 & 1.13 & 0.80 & 0.76 & 0.46 & 0.77 & \textbf{0.19} \\
    \bottomrule
    \end{tabular}
\end{table*}

\end{document}